%% file: main.tex
\documentclass{article}

\usepackage{
    algorithm,algorithmic,amsfonts,amsmath,amssymb,array,bm,booktabs,color,enumerate,graphicx,
    hyperref,makecell,microtype,multicol,natbib,nicefrac,subcaption,times,url,wrapfig
}
\usepackage[utf8]{inputenc}
\usepackage[T1]{fontenc}

\usepackage[accepted]{icml2019}

\include{defs}

\icmltitlerunning{\metabbr: Deep Structured Representations for Model-Based Reinforcement Learning}

\begin{document}

\twocolumn[
\icmltitle{\metabbr: Deep Structured Representations for\\Model-Based Reinforcement Learning}

\icmlsetsymbol{equal}{*}

\begin{icmlauthorlist}
\icmlauthor{Marvin Zhang}{equal,a}
\icmlauthor{Sharad Vikram}{equal,b}
\icmlauthor{Laura Smith}{a}
\icmlauthor{Pieter Abbeel}{a}
\icmlauthor{Matthew J.~Johnson}{c}
\icmlauthor{Sergey Levine}{a}
\end{icmlauthorlist}

\icmlaffiliation{a}{University of California, Berkeley}
\icmlaffiliation{b}{University of California, San Diego}
\icmlaffiliation{c}{Google}

\icmlcorrespondingauthor{Marvin Zhang}{marvin@eecs.berkeley.edu}

\icmlkeywords{}

\vskip 0.3in
]

\printAffiliationsAndNotice{\icmlEqualContribution}

\begin{abstract}

Model-based reinforcement learning (RL) has proven to be a data efficient approach for learning control tasks but is difficult to utilize in domains with complex observations such as images. In this paper, we present a method for learning representations that are suitable for iterative model-based policy improvement, even when the underlying dynamical system has complex dynamics and image observations, in that these representations are optimized for inferring simple dynamics and cost models given data from the current policy. This enables a model-based RL method based on the linear-quadratic regulator (LQR) to be used for systems with image observations. We evaluate our approach on a range of robotics tasks, including manipulation with a real-world robotic arm directly from images. We find that our method produces substantially better final performance than other model-based RL methods while being significantly more efficient than model-free RL.

\end{abstract}

\section{Introduction}
\label{sec:intro}

Model-based reinforcement learning (RL) methods use known or learned models in a variety of ways, such as planning through the model and generating synthetic experience \citep{dyna,rl-robotics-survey}. On simple, low-dimensional tasks, model-based approaches have demonstrated remarkable data efficiency, learning policies for systems like cart-pole swing-up with under 30 seconds of experience \citep{pilco,ode}. However, for more complex domains, one of the main difficulties in applying model-based methods is \emph{modeling bias}: if control or policy learning is performed against an imperfect model, performance in the real world will typically degrade with model inaccuracy \citep{pilco}. Many model-based methods rely on accurate forward prediction for planning \citep{nn-dyn,pets}, and for image-based domains, this precludes the use of simple models which will introduce significant modeling bias. However, complex, expressive models must typically be trained on very large datasets, corresponding to days to weeks of data collection, in order to generate accurate forward predictions of images \citep{dvf,l2g,l2p}.

\begin{figure}
    \centering
    \includegraphics[width=0.9\linewidth]{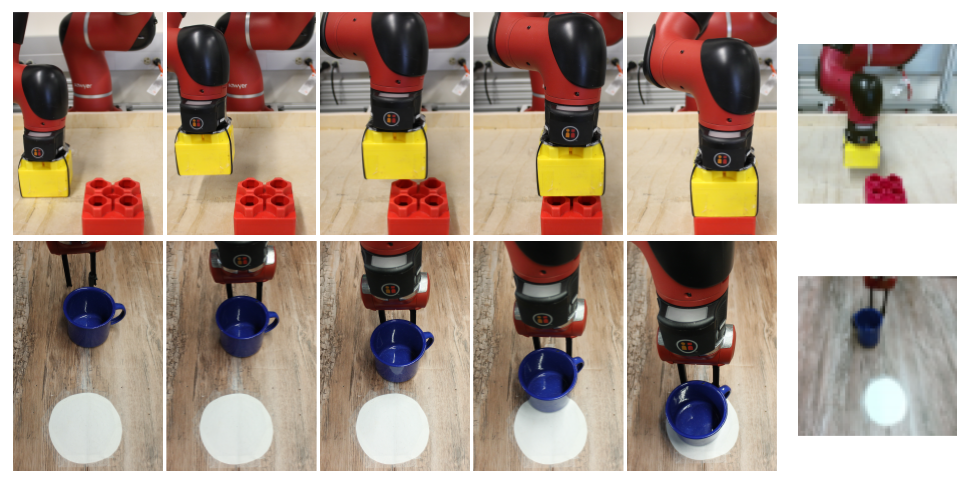}
    \caption{Our method can learn policies for complex manipulation tasks on a real Sawyer robot arm including stacking blocks (top) and pushing a mug onto a coaster (bottom), both from only \mbox{64-by-64-by-3} image observations (right), with no additional sensor information, and in one to two hours of interaction time.}
    \label{fig:diag}
\end{figure}

How can we use model-based methods to learn from images with similar data efficiency as we have seen in simpler domains? In our work, we focus on removing the need for accurate forward prediction, using what we term \emph{local models methods}. These methods use simple models, typically linear models, to provide gradient directions for local policy improvement, rather than for forward prediction and planning \citep{ilqg,mfcgps}. Thus, local model methods circumvent the need for accurate predictive models, but these methods cannot be directly applied to image-based tasks because image dynamics, even locally speaking, are highly non-linear.

Our main contribution is a representation learning and model-based RL procedure, which we term \methodname\ (\metabbr), that jointly optimizes a latent representation and model such that inference produces local models that provide good gradient directions for policy improvement. As shown in \autoref{fig:diag}, \metabbr\ is able to learn policies directly from high-dimensional image observations in several domains, including a real robotic arm stacking blocks and pushing objects with only one to two hours of data collection. To our knowledge, \metabbr\ is the most efficient RL method for solving real world robotics tasks directly from raw images. We also demonstrate several additional advantages of our method, including the ability to transfer learned models in the multi-task RL setting and the ability to handle sparse reward settings with a set of goal images.

\section{Preliminaries}
\label{sec:prelim}

We formalize our setting as a partially observed Markov decision process (POMDP) environment, which is given by the tuple $\MDP=(\Observations,\States,\Actions,\dynamics,\cost,\decoder,\initstatedist,\horizon)$. Most prior work in model-based RL assumes the fully observed RL setting where the observation space $\Observations$ is the same as the state space $\States$ and the observation density function $\decoder(\observation|\state)=\delta\{\observation=\state\}$ provides the exact state, so we will first discuss this setting. In this setting, the state space $\States$, action space $\Actions$, and horizon $\horizon$ are known, but the dynamics function $\dynamics(\state_{t+1}|\state_t,\action_t)$, cost function $\cost(\state_t,\action_t)$, and initial state distribution $\initstatedist(\state_1)$ are unknown. RL agents interact with the environment via a policy $\policy(\action_t|\state_t)$ that chooses an action conditioned on the current state, and the environment responds with the next state, sampled from the dynamics function, and the cost, evaluated through the cost function. The goal of RL is to minimize, with respect to the agent's policy, the expected sum of costs $\policyobj[\policy]=\mathbb{E}_{\policy,\dynamics,\initstatedist}\left[\sum_{t=1}^\horizon\cost(\state_t,\action_t)\right]$. Local model methods iteratively fit dynamics and cost models $\isdmodel,\dynmodel,\costmodel$ to data collected from the current policy in order to optimize $\hat\policyobj[\policy]\triangleq\mathbb{E}_{\policy,\dynmodel,\isdmodel}\left[\sum_{t=1}^\horizon\costmodel(\state_t,\action_t)\right]$. One particularly tractable and popular model is the linear-quadratic system (LQS), which models the dynamics as time-varying linear-Gaussian (TVLG) and the cost as quadratic, i.e.,
\begin{align*}
    \dynmodel(\state_{t+1}|\state_t,\action_t)&=\mathcal{N}\left(\state_{t+1}~\bigg|~\dynmat_t\colvec{\state_t\\\action_t},\dyncovar_t\right)\,,\\
    \costmodel(\state_t,\action_t)&=\frac{1}{2}\colvec{\state_t\\\action_t}^\top\costmat\colvec{\state_t\\\action_t}+\costvec^\top\colvec{\state_t\\\action_t}\,.
\end{align*}
Any deterministic policy operating in an environment with smooth dynamics can be locally modeled with a time-varying LQS \citep{convex}, while low-entropy stochastic policies are modeled approximately. This makes the time-varying LQS a reasonable local model for many dynamical systems. Furthermore, the optimal maximum-entropy policy $\policy^\star$ under the model is linear-Gaussian state feedback \citep{ddp}, i.e.,
\begin{align*}
    &\policy^\star(\action_t|\state_t)=\mathcal{N}\left(\K_t\state_t+\k_t,\polcovar_t\right)\,.
\end{align*}
We describe how to compute the parameters $\K_t$, $\k_t$, and $\polcovar_t$ in \autoref{sec:supp-lqr}. Due to modeling bias, the policy computed through LQR likely will not perform well in the real environment. This is because the model will not be globally correct but rather only valid close to the distribution of the data-collecting policy. One approach to addressing this issue is to use LQR with fitted linear models \citep[\mbox{LQR-FLM}; ][]{mfcgps}, a method which imposes a KL-divergence constraint on the policy update such that the shift in the trajectory distributions before and after the update, which we denote as $\bar{p}(\traj)$ and $p(\traj)$, respectively, is bounded by a step size $\polstepsize$. This leads to the constrained optimization
\begin{align}
    \max_\policy~\hat{\policyobj}[\policy]\text{\ \ \ s.t.\ \ \ }D_\KL(p(\traj)\|\bar{p}(\traj))\leq\polstepsize\,.\label{eq:lqrflm}
\end{align}
As shown in \citet{mfcgps}, this constrained optimization can be solved by augmenting the cost function to penalize the deviation from the previous policy $\bar{\policy}$, i.e., $\tilde{\cost}(\state_t,\action_t)=\frac{1}{\lambda}\costmodel(\state_t,\action_t)-\log\bar{\policy}(\action_t|\state_t)$. Note that this augmented cost function is still quadratic, since the policy is linear-Gaussian, and thus we can still compute the optimal policy for this cost function in closed form using the LQR procedure. $\lambda$ is a dual variable that trades off between optimizing the original cost and staying close in distribution to the previous policy, and the weight of this term can be determined through a dual gradient descent procedure.

Methods based on LQR have enjoyed considerable success in a number of control domains, including learning tasks on real robotic systems \citep{ilqg,gps}. However, most prior work in model-based RL assumes access to a low-dimensional state representation, and this precludes these methods from operating on complex observations such as images. There is some work on lifting this restriction: for example, \citet{e2c} and \citet{rce} combine LQR-based control with a representation learning scheme based on the variational auto-encoder \citep[VAE; ][]{vae-kingma,vae-rezende} where images are encoded into a learned low-dimensional representation that is used for modeling and control. They demonstrate success on learning several continuous control domains directly from pixel observations. We discuss our method's relationship to this work in \autoref{sec:related}.

\section{Learning and Modeling the Latent Space}
\label{sec:modeling}

\begin{figure}
    \centering
    \includegraphics[width=0.9\linewidth]{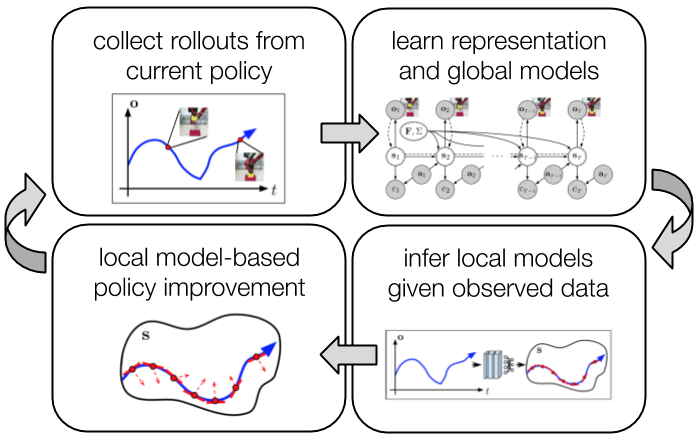}
    \caption{A high-level schematic of our method. We discuss the details of the model and inference procedure in \autoref{sec:modeling} and \autoref{sec:inference}. We then explain our algorithm in \autoref{sec:method}.}
    \label{fig:alg}
\end{figure}

Representation learning is a promising approach for integrating local models with complex observation spaces like images. What are the desired properties for a learned representation to be useful for local model methods? A simple answer is that local model fitting in a latent space that is low-dimensional and regularized will be more accurate than fitting directly to image observations. Concretely, one approach that satisfies these properties is to embed observations using a standard VAE, where regularization comes in the form of a unit Gaussian prior. However, a VAE representation still may not be amenable to local model fitting since the latent state is not optimized for dynamics and cost modeling. Since we aim to infer local dynamics and cost models in the neighborhood of the observed data, the main property we require from the latent representation is to make this fitting process more accurate for the observed trajectories, thereby reducing modeling bias and enabling a local model method to better improve the policy.

As we discuss in \autoref{sec:gen-model}, in order to make the local model fitting more accurate, especially in the low data regime, we learn global dynamics and cost models on all observed data jointly with the latent representation. Our formulation allows us to directly optimize the latent representation to be amenable for fitting linear dynamics and quadratic cost models, and \autoref{sec:var-family} details the learning procedure. \autoref{sec:inference} describes how, using our learned representation and global model as a starting point, we can infer local models that accurately explain the observed data. In this case, the local TVLG dynamics become latent variables in the model. As shown in \autoref{fig:alg}, updating the policy can then be done simply by rolling out a few trajectories, inferring the posterior over the latent TVLG dynamics, and using these dynamics and a local quadratic cost model to improve the policy. This procedure becomes the basis for the \metabbr\ algorithm which we present in \autoref{sec:method}.

\subsection{The Deep Bayesian LQS Model}
\label{sec:gen-model}

In our problem setting, we have access to trajectories of the form $\trajectory$ sampled from the system using our current policy. We assume this observed data is generated as follows: there is a latent state $\state$ that evolves according to linear-Gaussian dynamics, where the dynamics parameters themselves are stochastic and distributed according to a global prior. At each time step $t$, the latent state $\state_t$ is used to generate an image observation $\observation_t$, and the state and action generate the cost observation $\costsample_t$. The prior on the dynamics parameters increases the expressivity of the model by removing the assumption that the underlying dynamics are globally linear, since different trajectories may be explained by different samples from the prior. Furthermore, we approximate the observation function with a convolutional neural network, which makes the overall model non-linear. We formalize this generative model as
\begin{align}
    \state_1&\sim\N\left(0,\mathbf{I}\right)\,,\label{eq:gm-start}\\
    \dynmat,\dyncovar&\sim\mathcal{MNIW}\left(\Psi,\nu,\mathbf{M}_0,\mathbf{V}\right)\,,\label{eq:dyn}\\
    \state_{t+1}~|~\state_t,\action_t,\dynmat,\dyncovar&\sim\N\left(\dynmat\colvec{\state_t\\\action_t},\dyncovar\right)\,,\label{eq:dyn2}\\
    \observation_t~|~\state_t&\sim\decoder_\gamma\left(\state_t\right)\,,\\
    \costsample_t~|~\state_t,\action_t&\sim\N\left(\costmodel(\state_t,\action_t),1\right)\,.\label{eq:gm-end}
\end{align}
$\mathcal{MNIW}$ denotes the matrix normal inverse-Wishart (MNIW) distribution, which is the conjugate prior for linear-Gaussian dynamics models. Thus, conditioned on transitions from a particular time step, the posterior dynamics distribution $p\left(\dynmat,\dyncovar~\big|~\left\{\state^{(i)}_t,\action^{(i)}_t,\state^{(i)}_{t+1}\right\}_i\right)$ is still MNIW, and we describe in \autoref{sec:inference} how we leverage this conjugacy to infer local linear models using an approximate posterior distribution over the dynamics as a global prior. We refer to $\decoder_\gamma(\state)$ as an \emph{observation model} or \emph{decoder}, which is parameterized by neural network weights $\gamma$ and outputs a Bernoulli distribution over $\observation$, which are RGB images.

There are a number of ways to parameterize the quadratic cost model $\costmodel$, and we detail several options in \autoref{sec:supp-cost} along with an alternate parameterization for sparse human feedback that we discuss in \autoref{sec:method}.

\subsection{Joint Model and Representation Learning}
\label{sec:var-family}

We are interested in inferring two distributions of interest, both conditioned on the observations and actions:\footnote{Note that we do not condition on the cost observations for simplicity and also because the costs are scalars that contain relatively little information compared to image observations.}
\begin{enumerate}
    \itemsep0em
    \item The posterior distribution over dynamics parameters $p(\dynmat,\dyncovar~|~\observation_{1:\horizon},\action_{1:\horizon})$, as this informs our policy update;
    \item The posterior distribution over latent trajectories $p(\state_{1:\horizon}~|~\observation_{1:\horizon},\action_{1:\horizon},\dynmat,\dyncovar)$, since we require an estimate of the latent state as the input to our policy.
\end{enumerate}
The subscript $1:\horizon$ denotes an entire trajectory. Both of these distributions are intractable due to the neural network observation model. We instead turn to variational inference which optimizes, with respect to KL-divergence, a variational distribution $q$ in order to approximate a distribution of interest $p$. Specifically, we introduce the variational factors
\begin{align*}
    &q(\dynmat,\dyncovar)=\mathcal{MNIW}(\Psi',\nu',\bm{M}_0',\bm{V}')\,,\\
    &q\left(\state_{1:\horizon}~\big|~\dynmat,\dyncovar;\observation_{1:\horizon},\action_{1:\horizon}\right)\propto\\
    &\hspace{2em}p(\state_1)\prod_{t=1}^{\horizon-1}p\left(\state_{t+1}~\big|~\state_t,\action_t,\dynmat,\dyncovar\right)\prod_{t=1}^{\horizon}\psi(\state_t;\observation_t,\phi)\,.
\end{align*}
$q(\dynmat,\dyncovar)$ represents our posterior belief about the system dynamics after observing the collected data, and we also model this distribution as MNIW. We construct the full variational distribution over latent state trajectories as the normalized product of the state dynamics and, borrowing terminology from undirected graphical models, learned \emph{evidence potentials} $\psi(\state_t;\observation_t,\phi)=\N(\encoder_\phi(\observation_t))$. We refer to $\encoder_\phi(\observation)$ as a \emph{recognition model} or \emph{encoder}, which is parameterized by neural network weights $\phi$ and outputs the mean and diagonal covariance of a distribution over $\state$.

To learn the variational parameters, we optimize the evidence lower bound (ELBO), which is given by
\begin{align*}
    \mathcal{L}&=\mathbb{E}_q\left[\log\frac{p\left(\dynmat,\dyncovar,\state_{1:\horizon},\observation_{1:\horizon},\costsample_{1:\horizon}~\big|~\action_{1:\horizon}\right)}{q\left(\dynmat,\dyncovar,\state_{1:\horizon};\observation_{1:\horizon},\action_{1:\horizon}\right)}\right]\\
    &=\mathbb{E}_q\left[\sum_{t=1}^{\horizon}\log{p}(\observation_t|\state_t)\right]+\mathbb{E}_q\left[\sum_{t=1}^{\horizon}\log{p}(\costsample_t|\state_t,\action_t)\right]\\
    &\hspace{1em}-D_\KL\left(q\left(\dynmat,\dyncovar\right)\|p\left(\dynmat,\dyncovar\right)\right)\\
    &\hspace{1em}-\mathbb{E}_q\left[D_\KL\left(q\left(\state_{1:\horizon}~\big|~\dynmat,\dyncovar;\observation_{1:\horizon},\action_{1:\horizon}\right)\|\right.\right.\\
    &\hspace{6em}\left.\left.p\left(\state_{1:\horizon}~\big|~\action_{1:\horizon},\dynmat,\dyncovar\right)\right)\right]\,.
\end{align*}

\citet{svae} derived an algorithm for optimizing hybrid models with both deep neural networks and probabilistic graphical model (PGM) structure. In fact, our model bears strong resemblance to the LDS SVAE model from their work, though our ultimate goal is to fit local models for model-based policy learning rather than focusing on global models as in their work. We explain the relevant details of the SVAE learning procedure, which we use to learn the neural network parameters $\gamma$ and $\phi$ along with the global dynamics and cost models, in \autoref{sec:supp-svae}.

Note that, because the dynamics and cost are learned with samples from the recognition model, we backpropagate the gradients from the cost likelihood and dynamics KL terms through the encoder in order to learn a representation that is better suited to linear dynamics and quadratic cost. Through this, we learn a latent representation that, in addition to being low-dimensional and regularized, is directly optimized for fitting a LQS model on the observed data.

\begin{figure}
    \centering
    \includegraphics[width=0.9\linewidth]{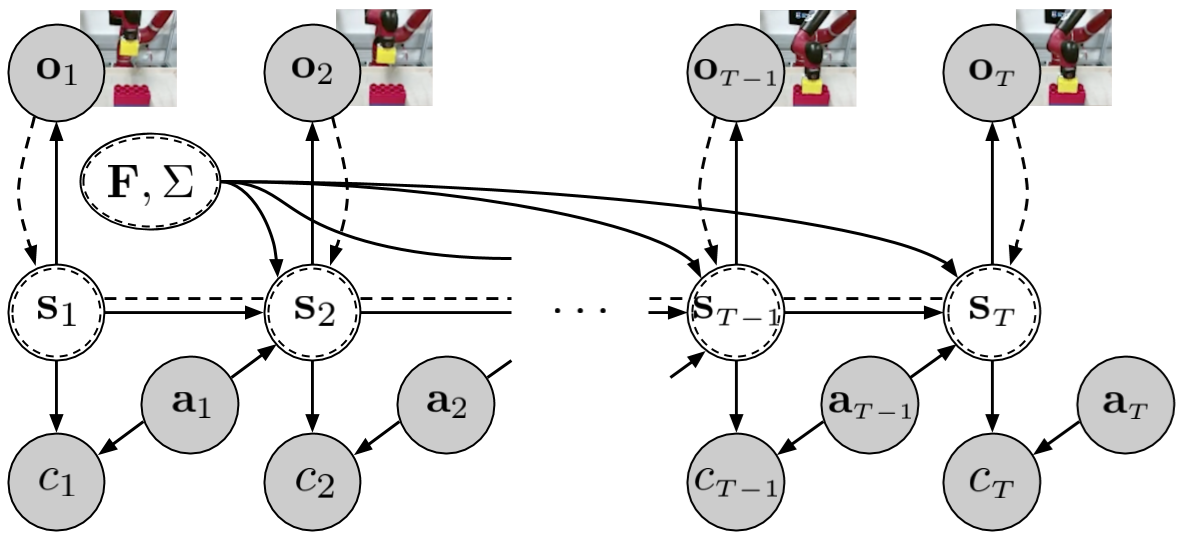}
    \caption{Our generative model, in solid lines, and variational family and recognition network, in dashed lines. In practice, the observations we work with are RGB images, and we use convolutional neural networks for both the recognition and observation models. The distributions for each node are specified in \autoref{sec:modeling}.}
    \label{fig:pgm}
\end{figure}

In \autoref{fig:pgm}, we depict our generative model using solid lines, and we depict the variational factors and recognition networks using dashed lines. Our method learns two variational distributions: first, a distribution over latent states which is used to provide inputs to the learned policy, and second, a global dynamics model that is used as a prior for inferring local linear dynamics models.

\section{Inference and RL in the Latent Space}
\label{sec:inference}

How can we utilize our learned representation and global models to enable local model methods? As shown in \autoref{fig:alg}, local model methods alternate between collecting batches of data from the current policy and using this data to fit local models and improve the policy. In order to improve the behavior of the local dynamics model fitting, especially in the low data regime, we use our global dynamics model as a prior and fit local dynamics models via posterior inference conditioned on data from the current policy.

For policy improvement, we fit local linear dynamics models separately at every time step, thus we augment the dynamics in our generative model from \autoref{eq:dyn} to instead be separate dynamics parameters $\dynmat_t,\dyncovar_t$ at each time step $t$. We model these parameters as independent samples from the global dynamics model $q(\dynmat,\dyncovar)$, and this can be interpreted as an empirical Bayes method, where we use data to estimate the parameters of our priors. In this way, the global dynamics model acts as a prior on the local time-varying dynamics models. In order to then infer the parameters of these local models conditioned on the data from the current policy, we employ a variational expectation-maximization (EM) procedure. The E-step computes $q\left(\state_{1:\horizon}~\big|~\dynmat_{1:\horizon},\dyncovar_{1:\horizon};\observation_{1:\horizon},\action_{1:\horizon}\right)$ given the current local dynamics, which are initialized to the global prior. The M-step optimizes, for each $t$, $\mathbb{E}[\log{q}(\dynmat_t,\dyncovar_t~\big|~\state_t,\action_t,\state_{t+1})]$ with respect to the dynamics parameters, where the expectation is over the latent state distribution from the E-step. We refer readers to \autoref{sec:supp-fit} for complete details.

We additionally fit a local quadratic cost model to the latest batch of data, and this combined with the local linear dynamics models gives us a local latent LQS model. Thus, it is natural to use LQR-based control in order to learn a policy. However, as discussed in \autoref{sec:prelim}, using vanilla LQR typically leads to undesirable behavior due to modeling bias.

One way to understand the problem is through standard supervised learning analysis, which only guarantees that our local models will be accurate under the distribution of data from the current policy. This directly motivates updating our policy in such a way that the trajectory distribution induced by the new policy does not deviate heavily from the data distribution, and in fact, the update rule proposed by LQR-FLM exactly accomplishes this goal \citep{mfcgps}. Thus, our policy update method utilizes the same constrained optimization from \autoref{eq:lqrflm}, and we solve this optimization using the same augmented cost function that penalizes deviation from the previous policy.

Note that rolling out our policy $\policy(\action_t|\state_t)$ requires computing an estimate of the current latent state $\state_t$. In order to handle partially observable tasks, we estimate the latent state using the history of observations and actions, i.e., $q\left(\state_t~\big|~\dynmat_{1:t-1},\dyncovar_{1:t-1};\observation_{1:t},\action_{1:t-1}\right)$, where we condition on the local linear dynamics fit to the latest batch of data. This distribution can be computed using Kalman filtering in the latent space and allows us to handle partial observability by aggregating information that may not be estimable from a single observation, such as system velocity from images.

\section{The \metabbr\ Algorithm}
\label{sec:method}

\begin{algorithm}[t]
\hspace*{\algorithmicindent}\textbf{Input:} \# iterations $K$; \# trajectories $\numsamp_{\text{init}},\numsamp$\\
\hspace*{\algorithmicindent}\textbf{Input:} model and policy hyperparameters $\xi_\model, \xi_\policy$\\
\hspace*{\algorithmicindent}\textbf{Output:} final model $\model$, final policy $\policy^{(K)}$\\
\begin{algorithmic}[1]
\STATE{$\policy^{(0)}\leftarrow\textsc{InitializePolicy}(\xi_\policy)$}
\STATE{$\dataset\leftarrow\textsc{CollectData}(\numsamp_{\text{init}},\policy^{(0)})$}
\STATE{$\model\leftarrow\textsc{TrainModel}(\dataset,\xi_\model)$ (\autoref{sec:modeling})}
\FOR{iteration $k\in\{1,\ldots,K\}$}
    \STATE{$\{\dynmat_t,\dyncovar_t\}_t\leftarrow\textsc{InferDynamics}(\dataset,\model)$ (\autoref{sec:inference})}
    \vspace{-1em}
    \STATE{$\policy^{(k)}\leftarrow\textsc{LQR-FLM}(\policy^{(k-1)},\{\dynmat_t,\dyncovar_t\}_t,\model)$\\(\autoref{sec:prelim})}
    \STATE{$\dataset\leftarrow\textsc{CollectData}(\numsamp,\policy^{(k)})$}
    \STATE{(optional) $\model\leftarrow\textsc{TrainModel}(\dataset,\xi_\model)$}
\ENDFOR
\end{algorithmic}
\caption{\metabbr}
\label{alg:method}
\end{algorithm}

The \metabbr\ algorithm is presented in \autoref{alg:method}. Lines~\mbox{1-3} detail the pretraining phase, corresponding to the representation and global model learning described in \autoref{sec:modeling}, where we collect $\numsamp_{\text{init}}$ trajectories using a random policy to train the representation, dynamics, and cost model. In our experiments in \autoref{sec:experiments}, we typically set $\numsamp_{\text{init}}\gg\numsamp$. In the RL phase, we alternate between inferring dynamics at each time step conditioned on data from the latest policy as described in \autoref{sec:inference} (line~5), performing the LQR-FLM update described in \autoref{sec:prelim} given the inferred dynamics (line~6), collecting $\numsamp$ trajectories using the updated policy (line~7), and optionally fine-tuning the model on the new data (line~8).\footnote{In our experiments, we found that fine-tuning the model did not improve final performance, though this step may be more important for environments where exploration is more difficult.} The model hyperparameters $\xi_\model$ include number of iterations, learning rates, and minibatch size, and the policy hyperparameters $\xi_\policy$ include the policy update KL constraint $\polstepsize$ and the initial random variance.

We evaluate \metabbr\ in \autoref{sec:experiments} in several RL settings involving continuous control including manipulation tasks on a real Sawyer robot. Beyond our method's performance on these tasks, however, we can derive several other significant advantages from our representation and PGM learning. As we detail in the rest of this section, these advantages include transfer in the multi-task RL setting and handling sparse reward settings using an augmented graphical model.

\subsection{Transferring Representations and Models}
\label{sec:transfer}

In the scenario where the dynamics are unknown, LQR-based methods are typically used in a ``trajectory-centric'' fashion where the distributions over initial conditions and goal conditions are low variance \citep{mfcgps,pilqr}. We similarly test our method in such settings in \autoref{sec:experiments}, e.g., learning Lego block stacking where the top block starts in a set position and the bottom block is fixed to the table. In the more general case where we may wish to handle several different conditions, we can learn a policy for each condition, however this may require significant amounts of data if there are many conditions.

However, one significant advantage of representation and model learning over alternative approaches, such as model-free RL, is the potential for transferring knowledge across multiple tasks where the underlying system dynamics do not change \citep{srl-survey}. Here, we consider each condition to be a separate task, and given a task distribution, we first sample various tasks and learn our model from \autoref{sec:modeling} using random data from these tasks. We show in \autoref{sec:experiments} that this ``base model'' can then be directly transferred to new tasks within the distribution, essentially removing the pretraining phase and dramatically speeding up learning for the Sawyer Lego block stacking domain.

\subsection{Learning from Sparse Rewards}
\label{sec:sparse}

Reward functions can often be hard to specify for complex tasks in the real world, and in particular they may require highly instrumented setups such as motion capture when operating from image observations. In these settings, sparse feedback is often easier to specify as it can come directly from a human labeler. Because we incorporate PGM machinery in our learned latent representation, it is straightforward for \metabbr\ to handle alternate forms of supervision simply by augmenting our generative model to reflect how the new supervision is given. Specifically, we extend our cost model to the sparse reward setting by assuming that we observe a binary signal $f_t$ based on the policy performance, rather than costs $\costsample_t$, and then modeling $f_t$ as a Bernoulli random variable with probability given by
\[
    p(f_t=1~|~\state_t,\action_t)\propto\exp\left\{-\costmodel(\state_t,\action_t)\right\}
\]
Concretely, in our experiments, $f_t$ is generated by a human that only provides $f_t=1$ when the task is solved. This setup is reminiscent of \citet{vice}, though our goal is not to classify expert data from policy data. Learning $\costmodel$ from observing $f_t$ amounts to logistic regression, and afterwards we can use $\costmodel$ as before in order to perform control and policy learning. Note that we can still backpropagate gradients through the encoder in order to learn a representation that is more amenable to predicting $f_t$. In \autoref{sec:experiments}, we use this method to solve a pushing task for which providing rewards is difficult without motion capture, and instead we use sparse human feedback and a set of goal images to specify the desired outcome. We provide the implementation details for this experiment in \autoref{sec:supp-set}.

\section{Related Work}
\label{sec:related}

Utilizing representation learning within model-based RL has been studied in a number of previous works \citep{srl-survey}, including using embeddings for state aggregation \citep{soft-state-aggregation}, dimensionality reduction \citep{dimension-reduction}, self-organizing maps \citep{self-organizing-map}, value prediction \citep{vpn}, and deep auto-encoders \citep{ae-nn,darla}. Among these works, deep spatial auto-encoders \citep[DSAE; ][]{spatial-ae} and embed to control \citep[E2C; ][]{e2c,rce} are the most closely related to our work, in that they consider local model methods combined with representation learning. The key difference in our work is that, rather than using a learning objective for reconstruction and forward prediction, our objective is more suited for local model methods by directly encouraging learning representations where fitting local models accurately explains the observed data. We also do not assume a known cost function, goal state, or access to the underlying system state as in DSAE and E2C, making \metabbr\ applicable even when the underlying states and cost function are unknown.\footnote{These methods may be extended to unknown underlying states and cost functions, though the authors do not experiment with this and it is unclear how well these approaches would generalize.}

Subsequent to our work, \citet{planet} formulate a representation and model learning method for image-based continuous control tasks that is used in conjunction with model-predictive control (MPC), which plans $H$ time steps ahead using the model, executes an action based on this plan, and then re-plans after receiving the next observation. We compare to a baseline that uses MPC in \autoref{sec:experiments}, and we empirically demonstrate the relative strengths of \metabbr\ and MPC, showing that \metabbr\ can overcome the short-horizon bias that afflicts MPC. We also compare to robust locally-linear controllable embedding \citep[RCE; ][]{rce}, an improved version of E2C, and we find that our approach tends to produce better empirical results.

\section{Experiments}
\label{sec:experiments}

We aim to answer the following through our experiments:
\begin{enumerate}
    \itemsep0em
    \item What benefits do we derive by utilizing model-based RL and representation learning in general?
    \item How does \metabbr\ compare to similar methods in terms of solving image-based control tasks?
    \item Can we utilize \metabbr\ to solve image-based control tasks on a real robotic system?
\end{enumerate}
To answer 1, we compare \metabbr\ to PPO \citep{ppo}, a state-of-the-art model-free RL method, and LQR-FLM with no representation learning. For the real world tasks, we also compare to deep visual foresight \citep[DVF; ][]{vf}, a state-of-the-art model-based method for images which does not use representation learning.

To answer 2, we compare to RCE \citep{rce}, which as discussed earlier is an improved version of E2C \citep{e2c}. We also set up an ``VAE ablation'' of \metabbr\ where we replace our representation learning scheme with a standard VAE. Finally, we consider an ``MPC baseline'' where we train neural network dynamics and cost models jointly with a latent representation and then use MPC with these models. Details regarding each of the comparisons are in \autoref{sec:comp}.

To answer 3, we evaluate \metabbr\ on a block stacking task and a pushing task on a Sawyer robot arm as shown in \autoref{fig:diag}. Videos of the learned policies are available at \mbox{\footnotesize{\url{https://sites.google.com/view/icml19solar}}.}

\subsection{Experimental Tasks}

\begin{figure}
    \centering
    \includegraphics[width=\linewidth]{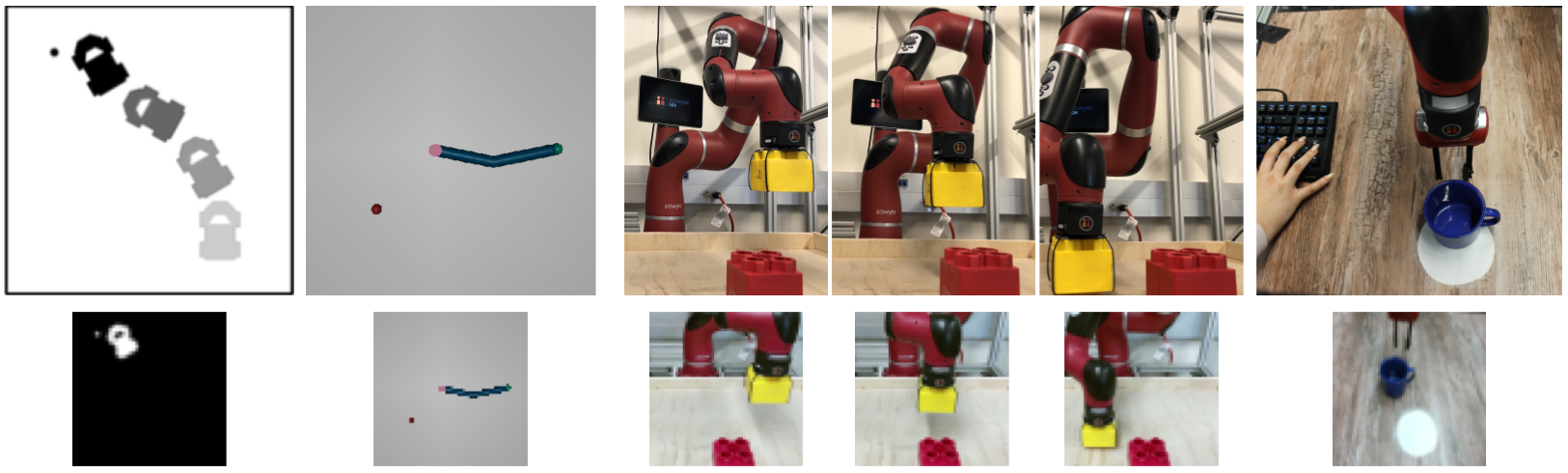}
    \caption{Illustrations of the environments we test on in the top row with example image observations in the bottow row. Left to right:~visualizing a trajectory in the nonholonomic car environment, with the target denoted by the black dot; an illustration of the 2-DoF reacher environment, with the target denoted by the red dot; the different tasks that we test for block stacking, where the rightmost task is the most difficult as the policy must learn to first lift the yellow block before stacking it; a depiction of our pushing setup, where a human provides the sparse reward that indicates whether the robot successfully pushed the mug onto the coaster.}
    \label{fig:envs}
\end{figure}

We set up simulated image-based robotic domains as well as manipulation tasks on a real Sawyer robotic arm, as shown in \autoref{fig:envs}. Details regarding task setup and training hyperparameters are provided in \autoref{sec:supp-set}.

{\bf 2D navigation.} Our 2-dimensional navigation task is similar to \citet{e2c} and \citet{rce} where an agent controls its velocity in a bounded planar system to reach a specified target. However, we make this task harder by randomizing the goal every episode rather than fixing it to the bottom right. Observations consist of two \mbox{32-by-32} images showing the positions of the agent and goal.

{\bf Nonholonomic car.} The nonholonomic car starts in the bottom right of the 2-dimensional space and controls its acceleration and steering velocity in order to reach the target in the top left. We use \mbox{64-by-64} images as the observation.

{\bf Reacher.} We experiment with the reacher environment from OpenAI Gym \citep{gym}, where a 2-DoF arm in a 2-dimensional plane has to reach a fixed target denoted by a red dot. For observations, we directly use \mbox{64-by-64-by-3} images of the rendered environment, which provides a top-down view of the reacher and target.

{\bf Sawyer Lego block stacking.} To demonstrate a challenging domain in the real world, we use our method to learn Lego block stacking with a real 7-DoF Sawyer robotic arm. The observations are \mbox{64-by-64-by-3} images from a camera pointed at the robot, and the controller only receives images as the observation without joint angles or other information. As shown in \autoref{fig:envs}, we define different block stacking tasks as different initial positions of the Sawyer arm.

{\bf Sawyer pushing.} We also experiment with the Sawyer arm learning to push a mug onto a white coaster, where we again use \mbox{64-by-64-by-3} images with no auxiliary information. Furthermore, we set up this task with only sparse binary rewards that indicate whether the mug is on top of the coaster, which are provided by a human labeler.

\subsection{Comparisons to Prior Work}

\begin{figure}
    \centering
    \begin{subfigure}{0.32\linewidth}
        \includegraphics[width=\linewidth]{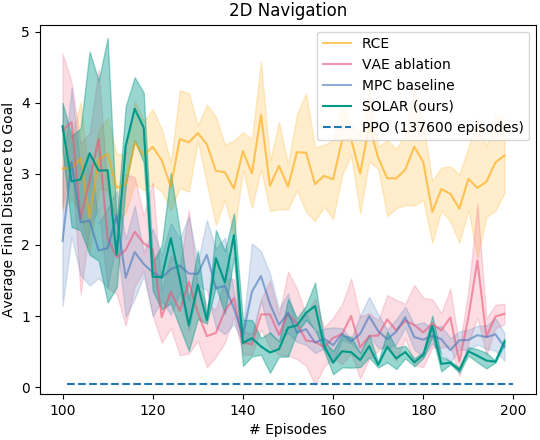}
        \caption{}
    \end{subfigure}
    \begin{subfigure}{0.32\linewidth}
        \includegraphics[width=\linewidth]{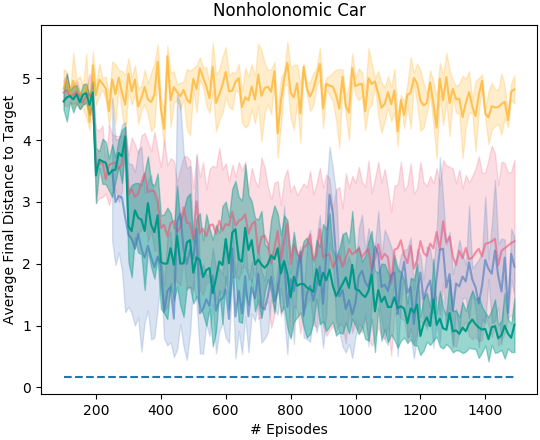}
        \caption{}
    \end{subfigure}
    \begin{subfigure}{0.33\linewidth}
        \includegraphics[width=\linewidth]{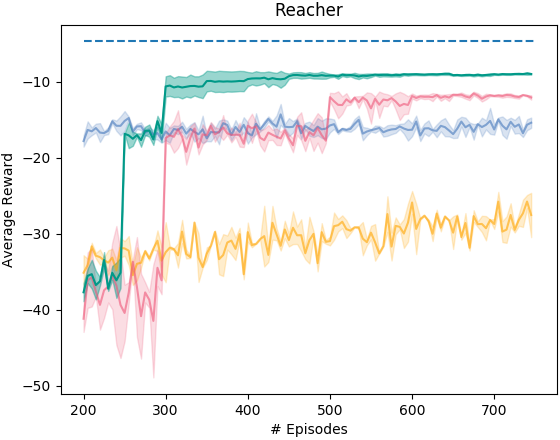}
        \caption{}
    \end{subfigure}
    \caption{Full size versions of these plots are available on the project website. (a):~Our method, the MPC baseline, and the VAE ablation consistently solve 2D navigation with a randomized goal, whereas RCE is unable to make progress. The final performance of PPO is plotted as the dashed line, though PPO requires 1000 times more samples than our method to reach this performance. (b):~On the nonholonomic car, both our method and the MPC baseline are able to reach the goal, though the VAE ablation is less consistent across seeds and RCE once again is unsuccessful at the task. PPO requires over 25 times more episodes than our method to learn a successful policy. (c):~On reacher, we perform worse than PPO but use about 40 times fewer episodes. RCE fails to learn at all, and the VAE ablation and MPC baseline are noticeably worse than our method. Here we plot reward, so higher is better.}
    \label{fig:sim-results}
\end{figure}

As shown in \autoref{fig:sim-results}, we compare to prior methods only on the simulated domains as these methods have not been shown to solve real-world image-based domains with reasonable data efficiency. On the 2D navigation task, our method, the VAE ablation, and the MPC baseline are able to learn very quickly, converging to high-performing policies in 200 episodes. However, these policies still exhibit some ``jittery'' behavior due to modeling bias, especially for the VAE ablation, whereas PPO learns an extremely accurate policy that continues to improve the longer we train. This gain in asymptotic performance is typical of model-free methods over model-based methods, however achieving this performance requires two to three orders of magnitude more samples. We present log-scale plots that illustrate the full learning progress of PPO in \autoref{sec:supp-exp}.

LQR-FLM from pixels fails to learn anything meaningful, and its performance does not improve over the initial policy. In fact, LQR-FLM does not make progress on any of the tasks, and for the sake of clarity in the plots, we omit these results. Similarly, despite extensive tuning and using code directly from the original authors, we were unable to get RCE to learn a good model for our 2D navigation task, and thus the learned policy also does not improve over the initial policy. RCE did not learn successful policies for any of the other tasks that we experiment with, though in \autoref{sec:supp-exp}, we show that RCE can indeed learn the easier fixed-target 2D navigation task from prior work.

On the nonholonomic car, our method and the MPC baseline are able to learn with about 1500 episodes of experience, whereas the VAE ablation's performance is less consistent. PPO eventually learns a successful policy for this task that performs better than our method, however it requires over 25 times more data to reach this performance.

Our method is outperformed by the final PPO policy on the reacher task, however, PPO requires about 40 times more data to learn. The VAE ablation and MPC baseline also make progress toward the target, though the performance is noticeably worse than our method. MPC often has better initial behavior than LQR-FLM as it uses the pretrained models right away for planning, highlighting one benefit of planning-based methods, however the MPC baseline barely improves past this behavior. Forward prediction with this learned model deteriorates quickly as the horizon increases, which makes long-horizon planning impossible. MPC is thus limited to short-horizon planning, and this limitation has been noted in prior work \citep{nn-dyn,mbve}. \metabbr\ does not suffer from this as we do not use our models for forward prediction.

Our open-source implementation of \metabbr\ is available at \mbox{\footnotesize{\url{https://github.com/sharadmv/parasol}}.}

\subsection{Analysis of Real Robot Results}

\begin{figure}
    \centering
    \begin{subfigure}{0.32\linewidth}
        \includegraphics[width=\linewidth]{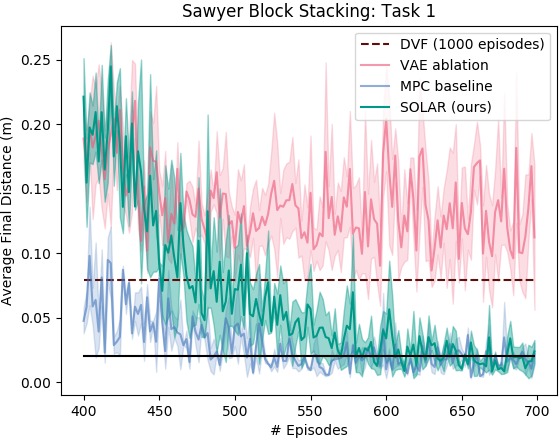}
        \caption{}
    \end{subfigure}
    \begin{subfigure}{0.32\linewidth}
        \includegraphics[width=\linewidth]{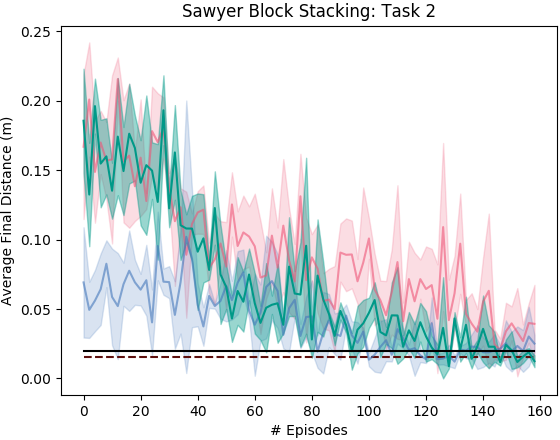}
        \caption{}
    \end{subfigure}
    \begin{subfigure}{0.32\linewidth}
        \includegraphics[width=\linewidth]{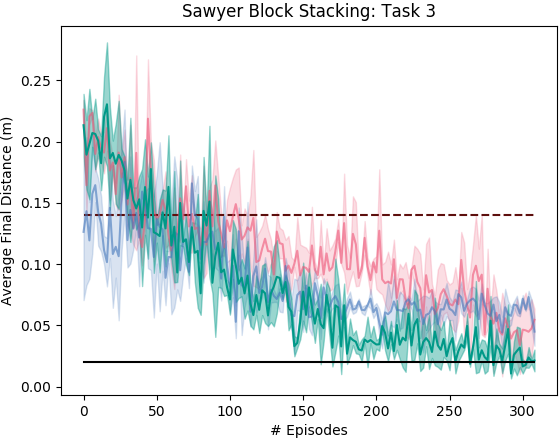}
        \caption{}
    \end{subfigure}
    \caption{Our method consistently solves all block stacking tasks. The MPC baseline learns very quickly on the two easier tasks since it can plan through the pretrained model, however, due to the short-horizon planning, it performs significantly worse on the hard task on the right where the block starts on the table. The VAE ablation performs well on the easiest task in the middle but is unsuccessful on the two harder tasks. DVF makes progress for each task but ultimately is not as data efficient as \metabbr. The black solid line at $0.02$m denotes successful stacking.}
    \label{fig:stacking-results}
\end{figure}

The real-world Lego block stacking results are shown in \autoref{fig:stacking-results}. Our method is successful on all tasks, where we define success as achieving an average distance of $0.02$m which generally corresponds to successful stacking, whereas the VAE ablation is only successful on the easiest task in the middle plot. The MPC baseline again starts off better and learns more quickly on the two easier tasks. However, MPC is again limited to short-horizon planning, which causes it to fail on the most difficult task in the right plot as it simply greedily reduces the distance between the two blocks rather than lifting the block off the table. We can solve each block stacking task using about two hours of robot interaction time, though the x-axes in the plots show that we further reduce the total data requirements by about a factor of two by pretraining and transferring a shared representation and global model as described in \autoref{sec:method}.

As a comparison to a state-of-the-art model-based method that has been successful in real-world image-based domains, we evaluate DVF \citep{vf}, which learns pixel space models and does not utilize representation learning. We find that this method can make progress but ultimately is not able to solve the two harder tasks even with more data than what we use for our method and even with a much smaller model. This highlights our method's data efficiency, as we use about two hours of robot data compared to days or weeks of data as in this prior work.

\begin{table}
    \centering
    \begin{tabular}{c||c|c}
        \hline
        \textbf{Method} & \thead{Final Distance\\to Goal (cm)} & \thead{Episodes\\per Seed} \\
        \hline
        \makecell{DVF\\\citep{vf}} & $4.50\pm2.60$ & $280$ \\
        \hline
        \metabbr\ (ours) & $1.85\pm0.86$ & $250$ \\
    \end{tabular}
    \caption{Sawyer Pushing with Sparse Rewards}
    \label{tab:pushing-results}
\end{table}

\begin{figure}
    \centering
    \includegraphics[width=\linewidth]{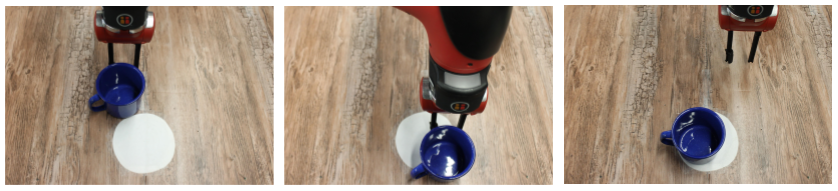}
    \caption{Visualizing example end states from rolling out our policy after 200 (top), 230 (middle) and 260 (bottom) trajectories.}
    \label{fig:pushing-viz}
\end{figure}

Finally, on the real-world pushing task, despite the additional challenge of sparse rewards, our method learns a successful policy in about an hour of interaction time as detailed in \autoref{tab:pushing-results} and visualized in \autoref{fig:pushing-viz}. DVF performs worse than our method with a comparable amount of data, again even when using a downsized model. Videos depicting the learning process for both of the real-world tasks, as well as full size versions of the plots and learning curves, are available at \mbox{\footnotesize{\url{https://sites.google.com/view/icml19solar}}.}

\section{Discussion}
\label{sec:discussion}

We presented \metabbr, a model-based RL algorithm that is capable of learning policies in a data-efficient manner directly from raw high-dimensional image observations. The key insights in \metabbr\ involve learning latent representations where simple models are more accurate and utilizing PGM structure to infer dynamics from data conditioned on observed trajectories. Our experimental results demonstrate that \metabbr\ is competitive in sample efficiency, while exhibiting superior final policy performance, compared to other model-based methods. \metabbr\ is also significantly more data-efficient compared to model-free RL methods, especially when transferring previously learned representations and models. We show that \metabbr\ can learn complex real-world robotic manipulation tasks with only image observations in one to two hours of interaction time.

Our model is designed for and tested on continuous action domains, and extending our model to discrete actions would necessitate some type of learned action representation. This is intriguing also as a potential mechanism for further reducing modeling bias. Certain systems such as dexterous hands and tensegrity robots not only exhibit complex state spaces but also complex action spaces \citep{hand-rail,hand-openai,superball}, and learning simpler action representations that can potentially capture high-level behavior, such as manipulation or locomotion primitives, is an exciting line of future work.

\paragraph*{Acknowledgments.} MZ is supported by an NDSEG fellowship. SV is supported by NSF grant CNS1446912. This work was supported by the NSF, through IIS-1614653, and computational resource donations from Amazon.

\bibliography{references}
\bibliographystyle{icml2019}

\appendix
\include{appendix}

\end{document}

%% file: defs.tex
\newcommand{\MDP}{M}
\newcommand{\Observations}{\mathcal{O}}
\newcommand{\States}{\mathcal{S}}
\newcommand{\Actions}{\mathcal{A}}
\newcommand{\cost}{C}
\newcommand{\initstatedist}{\rho}

\newcommand{\numsamp}{N}
\newcommand{\horizon}{T}
\newcommand{\dynamics}{p}

\newcommand{\policy}{\pi}
\newcommand{\state}{\mathbf{s}}
\newcommand{\action}{\mathbf{a}}
\newcommand{\costsample}{c}
\newcommand{\policyobj}{\eta}
\newcommand{\dynmodel}{\hat{\dynamics}}
\newcommand{\costmodel}{\hat{\cost}}
\newcommand{\isdmodel}{\hat{\rho}}
\newcommand{\dynmat}{\mathbf{F}}
\newcommand{\dyncovar}{\Sigma}
\newcommand{\costmat}{\mathbf{C}}
\newcommand{\costvec}{\mathbf{c}}
\newcommand{\K}{\mathbf{K}}
\renewcommand{\k}{\mathbf{k}}
\newcommand{\polcovar}{\mathbf{S}}
\newcommand{\polstepsize}{\epsilon}
\newcommand{\observation}{\mathbf{o}}
\newcommand{\encoder}{e}
\newcommand{\decoder}{f}
\newcommand{\traj}{\tau}
\newcommand{\KL}{\text{KL}}
\newcommand{\N}{\mathcal{N}}
\newcommand{\model}{\mathcal{M}}
\newcommand{\dataset}{\mathcal{D}}
\newcommand{\colvec}[2][1]{%
  \scalebox{#1}{%
    \renewcommand{\arraystretch}{#1}%
    $\begin{bmatrix}#2\end{bmatrix}$%
  }
}

\newcommand{\trajectory}{\left[\observation_0,\action_0,\costsample_0,\ldots,\observation_{\horizon},\action_{\horizon},\costsample_{\horizon}\right]}

\newcommand{\methodname}{stochastic optimal control with latent representations}
\newcommand{\metabbr}{SOLAR}

%% file: appendix.tex
\section{Policy Learning Details}
\label{sec:supp-lqr}

Given a TVLG dynamics model and quadratic cost approximation, we can approximate our Q and value functions to second order with the following dynamic programming updates, which proceed from the last time step $t = T$ to the first step $t = 1$:
\begin{align*}
    Q_{\mathbf{s},t}=c_{\mathbf{s},t}&+\dynmat_{\mathbf{s},t}^\top{V}_{\mathbf{s},t+1}\,,~~Q_{\mathbf{ss},t}=c_{\mathbf{ss},t}+\dynmat_{\mathbf{s},t}^\top{V}_{\mathbf{ss},t+1}\dynmat_{\mathbf{s},t}\,,\\
    Q_{\mathbf{a},t}=c_{\mathbf{a},t}&+\dynmat_{\mathbf{a},t}^\top{V}_{\mathbf{s},t+1}\,,~~Q_{\mathbf{aa},t}=c_{\mathbf{aa},t}+\dynmat_{\mathbf{a},t}^\top{V}_{\mathbf{ss},t+1}\dynmat_{\mathbf{a},t}\,,\\
    &Q_{\mathbf{sa},t}=c_{\mathbf{sa},t}+\dynmat_{\mathbf{s},t}^\top{V}_{\mathbf{ss},t+1}\dynmat_{\mathbf{a},t}\,,\\
    &V_{\mathbf{s},t}=Q_{\mathbf{s},t}-Q_{\mathbf{sa},t}Q_{\mathbf{aa},t}^{-1}Q_{\mathbf{a},t}\,,\\
    &V_{\mathbf{ss},t}=Q_{\mathbf{ss},t}-Q_{\mathbf{sa},t}Q_{\mathbf{aa},t}^{-1}Q_{\mathbf{as},t}\,.
\end{align*}
It can be shown (e.g., by \citet{synthesis}) that the action $\mathbf{a}_t$ that minimizes the second-order approximation of the Q-function at every time step $t$ is given by
\[
\mathbf{a}_t=-Q_{\mathbf{aa},t}^{-1}Q_{\mathbf{as},t}\mathbf{s}_t-Q_{\mathbf{aa},t}^{-1}Q_{\mathbf{a},t}\,.
\]
This action is a linear function of the state $\mathbf{s}_t$, thus we can construct an optimal linear policy by setting $\mathbf{K}_t=-Q_{\mathbf{aa},t}^{-1}Q_{\mathbf{as},t}$ and $\mathbf{k}_t=-Q_{\mathbf{aa},t}^{-1}Q_{\mathbf{a},t}$. We can also show that the maximum-entropy policy that minimizes the approximate Q-function is given by
\[
\pi^\star(\mathbf{a}_t|\mathbf{s}_t)=\mathcal{N}(\mathbf{K}_t\mathbf{s}_t+\mathbf{k}_t,Q_{\mathbf{aa},t}).
\]
Furthermore, as in \citet{mfcgps}, we can impose a constraint on the total KL-divergence between the old and new trajectory distributions induced by the policies through an augmented cost function $\tilde{\cost}(\state_t,\action_t)=\frac{1}{\lambda}\costmodel(\state_t,\action_t)-\log\bar{\policy}(\action_t|\state_t)$, where solving for $\lambda$ via dual gradient descent can yield an exact solution to a KL-constrained LQR problem.

\section{Parameterizing the Cost Model}
\label{sec:supp-cost}

The simplest choice that we consider for parameterizing the cost model is as a full quadratic function of the state and action, i.e.,     $\costmodel(\state_t,\action_t)=\frac{1}{2}\state_t^\top\costmat\state_t+\costvec^\top\state_t+\alpha\|\action_t\|_2^2+b$ where we assume that the action-dependent part of the cost -- i.e., $\alpha$ -- is known, and we impose no restrictions on the learned parameters $\costmat$ and $\costvec$. This is our default option due to its simplicity and the added benefit that fitting this model locally can be done in closed form through least-squares quadratic regression on the observed states. However, another option we consider is to choose $\costmodel(\state_t,\action_t)=\frac{1}{2}\state_t^\top\mathbf{L}\mathbf{L}^\top\state_t+\costvec^\top\state_t+\alpha\|\action_t\|_2^2+b$. $\mathbf{L}$ is a lower-triangular matrix with non-negative diagonal entries, and thus by constructing our cost matrix as $\costmat=\mathbf{L}\mathbf{L}^\top$ we guarantee that the learned cost matrix is positive semidefinite, which can improve the behavior of the policy update.

In general, in this work, we consider quadratic parameterizations of the cost model since we wish to build a LQS model. However, in general it may be possible to use non-quadratic but twice-differentiable cost models, such as a neural network model, and compute local quadratic cost models using a second-order Taylor approximation as in \citet{mfcgps}. We also do not assume access to a goal observation, though if provided with such information we can construct a quadratic cost function that penalizes distance to this goal in the learned latent space, as in \citet{spatial-ae} and \citet{e2c}.

\section{The SVAE Algorithm}
\label{sec:supp-svae}

\citet{svae} build off of \citet{svi} and \citet{vmp}, who show that, for conjugate exponential models, the variational model parameters can be updated using natural gradients of the form
\begin{equation}
    \tilde{\nabla}_{\omega}\mathcal{L}=\omega^0+B\mathbb{E}_q\left[t_{\dynmat,\dyncovar}(\dynmat,\dyncovar)\right]-\omega\,,
\end{equation}
Where $\omega$ denotes the MNIW parameters of the variational factors on $\dynmat,\dyncovar$, $B$ is the number of minibatches in the dataset, $\omega^0$ is the parameter for the prior distribution $p(\dynmat,\dyncovar)$, and $t_{\dynmat,\dyncovar}(\dynmat,\dyncovar)$ is the sufficient statistic function for $p(\dynmat,\dyncovar)$. Thus, we can use this equation to compute the natural gradient update for $\omega$, whereas for $\gamma$, $\phi$, and the parameters of the cost model, we use stochastic gradient updates on Monte Carlo estimates of the ELBO, specifically using the Adam optimizer \citep{adam}. This leads to two simultaneous optimizations, and their learning rates are treated as separate hyperparameters. We have found $10^{-4}$ and $10^{-3}$ to be good default settings for the natural gradient step size and stochastic gradient step size, respectively.

\section{Fitting the Local Dynamics Model}
\label{sec:supp-fit}

In the pretraining phase described in \autoref{sec:modeling}, we are learning the following sets of parameters from observed trajectories:
\vspace{-.5em}
\begin{enumerate}
    \itemsep0em
    \item The parameters of the variational posterior over global dynamics $q_\textrm{global}(\dynmat, \dyncovar)$;
    \item The weights of the encoder and decoder networks $\decoder_\gamma(\state)$ and $\encoder_\phi(\observation)$;
    \item The parameters of the cost function $\costmodel(\state, \action)$.
\end{enumerate}
\vspace{-.5em}
In the RL phase described in \autoref{sec:inference}, after learning the representation and global models, we fit local linear-Gaussian dynamics models to additional trajectories. The conjugacy of the Bayesian LQS model enables a computationally efficient expectation-maximization procedure to learn the local dynamics. We assume the same graphical model as in \autoref{eq:gm-start} to \autoref{eq:gm-end} except we modify \autoref{eq:dyn} and \autoref{eq:dyn2} to be
\begin{align*}
    \dynmat_t, \dyncovar_t &\sim p(\dynmat_t, \dyncovar_t) \triangleq q_{\textrm{global}}(\dynmat, \dyncovar)\,,\\
    \state_{t + 1} | \state_t, \action_t, \dynmat_t, \dyncovar_t &\sim \N\left(\dynmat_t\colvec{\state_t\\\action_t}, \dyncovar_t\right)\,.
\end{align*}
The model assumes that the TVLG dynamics are independent samples from our global dynamics, followed by a deep Bayesian LDS to generate trajectories. This is similar to the globally trained model, with the exception that we explicitly assume time-varying dynamics.

Now suppose we have collected a set of trajectories of the form $\trajectory$ and aim to fit a local dynamics model. We use variational inference to approximate the posterior distributions by setting up the variational factors
\vspace{-.5em}
\begin{enumerate}
    \itemsep0em
    \item $q(\state_{1:\horizon} | \dynmat_{1:\horizon}, \dyncovar_{1:\horizon}; \observation_{1:\horizon}, \action_{1:\horizon})$, which approximates the posterior distribution $p(\state_{1:\horizon} | \observation_{1:\horizon}, \action_{1:\horizon}, \dynmat_{1:\horizon}, \dyncovar_{1:\horizon})$;
    \item $q(\dynmat_t, \dyncovar_t)$, which approximates the posterior distribution $p(\dynmat_t, \dyncovar_t | \state_{1:\horizon}, \action_{1:\horizon})$
\end{enumerate}
\vspace{-.5em}
The ELBO under these variational factors is:
\begin{align*}
    \mathcal{L} &= \mathbb{E}_q \big[\sum_t^\horizon \log p(\observation_t | \state_t)\\ &- \KL\left(q(\state_{1:\horizon})\|p(\state_{1:\horizon} | \action_{1:\horizon}, \dynmat_{1:\horizon}, \dyncovar_{1:\horizon})\right)\\
    &- \sum_t^{\horizon - 1} \KL\left(q(\dynmat_t, \dyncovar_t)\|p(\dynmat_t, \dyncovar_t)\right)
    \big]
\end{align*}

We use variational EM to alternatively optimize $q(\state_{1:\horizon} | \dynmat_{1:\horizon}, \dyncovar_{1:\horizon}; \observation_{1:\horizon}, \action_{1:\horizon})$ and $q(\dynmat_t, \dyncovar_t)$. Using evidence potentials $\psi(\state_t;\observation_t,\phi)$ output by the recognition network $\encoder_\phi(\observation_t)$, both of these optimizations can be done in closed form. Specifically, the optimal $q(\state_{1:\horizon} | \dynmat_{1:\horizon}, \dyncovar_{1:\horizon}; \observation_{1:\horizon}, \action_{1:\horizon})$ is computed via Kalman smoothing using evidence potentials from the recognition network, and the optimal $q(\dynmat_t, \dyncovar_t)$ can be computed via Bayesian linear regression using expected sufficient statistics from $q(\state_{1:\horizon} | \dynmat_{1:\horizon}, \dyncovar_{1:\horizon}; \observation_{1:\horizon}, \action_{1:\horizon})$.

\section{Experiment Setup}
\label{sec:supp-set}

{\bf 2D navigation.} Our recognition model architecture for the 2D navigation domain consists of two convolution layers with \mbox{2-by-2} filters and 32 channels each, with no pooling layers and ReLU non-linearities, followed by another convolution with \mbox{2-by-2} filters and 2 channels. The output of the last convolution layer is fed into a fully-connected layer which then outputs a Gaussian distribution with diagonal covariance. Our observation model consists of FC hidden layers with 256 ReLU activations, and the last layer outputs a categorical distribution over pixels. We initially collect 100 episodes which we use to train our model, and for every subsequent RL iteration we collect 10 episodes. The cost function we use is the sum of the $L^2$-norm squared of the distance to the target and the commanded action, with weights of 1 and 0.001, respectively.

As discussed in \autoref{sec:experiments}, we modify the 2D navigation task from \citet{e2c} and \citet{rce} to randomize the location of the target every episode, and we set this location uniformly at random between $-2.8$ and $2.8$ for both the x and y coordinates, as coordinates outside of $[-3,3]$ are not visible in the image. We similarly randomize the initial position of the agent. In this setup, we use two \mbox{32-by-32} images as the observation, one with the location of the agent and the other with the location of the target, and in the fixed-target version of the task we only use one \mbox{32-by-32} image.

{\bf Nonholonomic car.} The nonholonomic car domain consists of \mbox{64-by-64} image observations. Our recognition model is a convolutional neural network with four convolutional layers with \mbox{4-by-4} filters with 4 channels each, and the first two convolution layers are followed by a ReLU non-linearity. The output of the last convolutional layer is fed into three FC ReLU layers of width 2048, 512, and 128, respectively. Our final layer outputs a Gaussian distribution with dimension 8. Our observation model consists of four FC ReLU layers of width 256, 512, 1024, and 2048, respectively, followed by a Bernoulli distribution layer that models the image. For this domain, we collect 100 episodes initially to train our model, and then for RL we collect 100 episodes per iteration. The cost function we use is the sum of the $L^2$-norm squared of the distance from the center of the car to the target and the commanded action, with weights of 1 and 0.001, respectively.

{\bf Reacher.} The reacher domain consists of \mbox{64-by-64-by-3} image observations. Our recognition model consists of three convolutional layers with \mbox{7-by-7}, \mbox{5-by-5}, and \mbox{3-by-3} filters with 64, 32 and 8 channels respectively. The first convolutional layer is followed by a ReLU non-linearity. The output of the last convolutional layer is fed into an FC ReLU layer of width 256, which outputs a Gaussian distribution with dimension 10. Our observation model consists of one FC ReLU layers of width 512, followed by three deconvolutional layers with the reverse order of filters and channels as the recognition model. This is followed by a Bernoulli distribution layer that models each image. We collect 200 episodes initially to train our model, and then for RL we collect 100 episodes per iteration. The cost function we use is the sum of the $L^2$-norm of the distance from the fingertip to the target and the $L^2$-norm squared of the commanded action, which is the negative of the reward function as defined in Gym.

{\bf Sawyer Lego block stacking.} The image-based Sawyer block-stacking domain consists of \mbox{64-by-64-by-3} image observations. The policy outputs velocities on the end effector in order to control the robot. Our recognition model is a convolutional neural network with the following architecture: a \mbox{5-by-5} filter convolutional layer with 16 channels followed by two convolutional layers using \mbox{5-by-5} filters with 32 channels each. The convolutional layers are followed by ReLU activations leading to a 12 dimensional Gaussian distribution layer. Our observation model consists of a FC ReLU layer of width 128 feeding into three deconvolutional layers, the first with \mbox{5-by-5} filters with 16 channels and the last two of \mbox{6-by-6} filters with 8 channels each. These are followed by a final Bernoulli distribution layer.

For this domain, we collect 400 episodes initially to train our model and 10 per iteration thereafter. Note that this pretraining data is collected only once across solving all of the tasks that we test on. The cost function is the cubed root of the $L^2$-norm of the displacement vector between the end-effector and the target in 3D-space.

{\bf Sawyer pushing.} The image-based Sawyer pushing domain also operates on \mbox{64-by-64-by-3} image observations. Our recognition and observation models are the same as those used in the block-stacking domain. The dynamics model is learned by a network with two FC ReLU layers of width 128 followed by a 12 dimensional Gaussian distribution layer. The cost model is learned jointly with the representation and dynamics by optimizing the ELBO, which with regards to the cost corresponds to logistic regression on the observed sparse reward using a sampled latent state as the input. We collect 200 episodes to train our model and 20 per iteration for RL.

During the RL phase, the human supervisor uses keyboard input to provide the sparse reward signal to the learning algorithm, indicating whether or not the mug was successfully pushed onto the coaster. In practice, for simplicity, we label the last five images of the trajectory as either $0$ or $1$ depending on whether or not the keyboard was pressed at any time during the trajectory, as for this task a successful push is typically reflected in the end state. In order to overcome the exploration problem and provide a diverse dataset for pretraining the cost model, we manually collect $180$ ``goal images'' where the mug is on the coaster and the robot arm is in various locations.

\section{Implementation of Comparisons}
\label{sec:comp}

{\bf PPO.} We use the open source implementation of PPO (named ``PPO2'') from the OpenAI Baselines project: \mbox{\footnotesize{\url{https://github.com/openai/baselines}}}. We write OpenAI gym wrappers for our simulated environments in order to test PPO on our simulated tasks.

{\bf LQR-FLM.} We implement LQR-FLM based on the open-source implementation from the Guided Policy Search project: \mbox{\footnotesize{\url{https://github.com/cbfinn/gps}}}. The only modification to the LQR-FLM algorithm that we make is to handle unknown cost functions by fitting a quadratic cost model to data from the current policy.

{\bf DVF.} We train a video prediction model using the open source Stochastic Adversarial Video Prediction project: \mbox{\footnotesize{\url{https://github.com/alexlee-gk/video_prediction}}}. To define the task, we specify the location of a pixel whose movement to a specified goal location indicates success. The cost function is then the predicted probability of successfully moving the selected pixel to the goal. We then use MPC, specifically the cross-entropy method (CEM) for offline planning: we sample sequences of actions from a Gaussian, predict the corresponding sequence of images using the video prediction model, evaluate the cost of the imagined trajectory with the cost model, and refit the parameters of the Gaussian to the best predicted action sequences. This iterative process eventually outputs an action sequence to perform in the real world in order to try and solve the task.

{\bf RCE.} We use model learning code directly from the authors of RCE \citep{rce}, though this code is not publicly available and to our knowledge there are no open source implementations of RCE or E2C \citep{e2c} that are able to reproduce the results from the respective papers. In addition to LQR-based control, we also experiment with MPC with neural network dynamics and cost models in the learned latent representation. In our experiments, we report the best results using either of these control methods.

{\bf VAE ablation.} In the VAE ablation, we replace our representation and global models with a standard VAE \citep{vae-kingma,vae-rezende}, which imposes a unit Gaussian prior on the latent representation. Because we cannot infer local dynamics as described in \autoref{sec:inference}, we instead use a GMM dynamics prior that is trained on all data as described by \citet{gps}. After fitting a local quadratic cost model, we again have a local LQS model that we can use in conjunction with an LQR-FLM policy update.

{\bf MPC baseline.} (MPC) involves planning $H$ time steps ahead using a dynamics and cost model, executing an action based on this plan, and then re-planning after receiving the next observation \citep{mpc}. Recently, MPC has proven to be a successful control method when combined with neural network dynamics models, where many trajectories are sampled using the model and then the first action corresponding to the best imagined trajectory is executed \citep{nn-dyn,pets}. Similar to LQR-FLM, we can extend MPC to handle image-based domains by learning dynamics and cost models within a learned latent representation. As MPC does not require an LQS model, we can instead utilize neural network dynamics and cost models which are more expressive.

\section{Additional Experiments}
\label{sec:supp-exp}

\subsection{RCE on Fixed-Target 2D Navigation}

\begin{figure}
    \centering
    \includegraphics[width=0.8\linewidth]{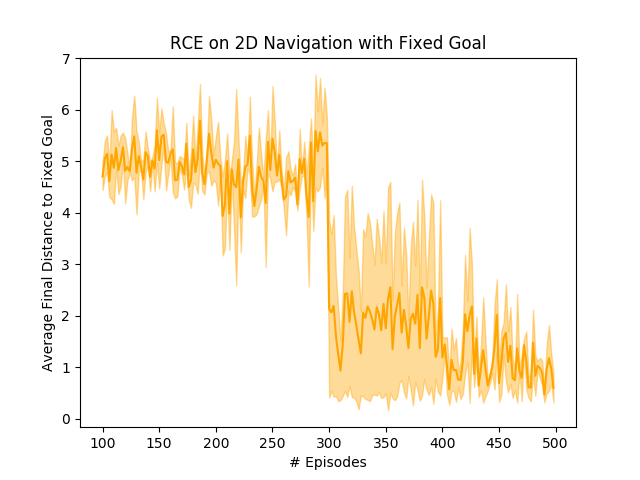}
    \caption{On 2D navigation with the goal fixed to the bottom right, RCE is able to successfully learn a policy for navigating to the goal.}
    \label{fig:pm-e2c}
\end{figure}

As mentioned in \autoref{sec:experiments}, RCE was unable to make progress for the 2D navigation task, though we were able to get more successful results by fixing the position of the goal to the bottom right as is done in the image-based 2D navigation task considered in E2C \citep{e2c} and RCE \citep{rce}. \autoref{fig:pm-e2c} details this experiment, which we ran for three random seeds and report the mean and standard deviation of the average final distance to the goal as a function of the number of training episodes. This indicates that RCE can indeed solve some tasks from image observations, though we were unable to use RCE succesfully on any of the tasks we consider.

\subsection{Full Learning Progress of PPO}

\begin{figure}
    \centering
    \begin{subfigure}{0.32\linewidth}
        \includegraphics[width=0.9\linewidth]{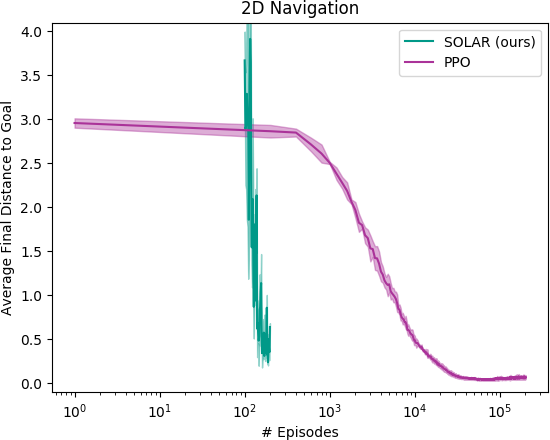}
        \caption{}
    \end{subfigure}
    \begin{subfigure}{0.31\linewidth}
        \includegraphics[width=0.9\linewidth]{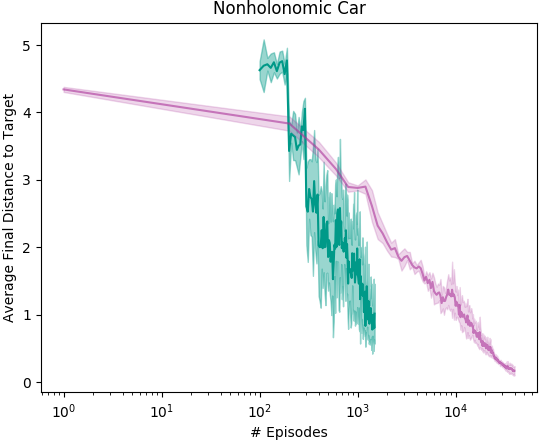}
        \caption{}
    \end{subfigure}
    \begin{subfigure}{0.33\linewidth}
        \includegraphics[width=0.9\linewidth]{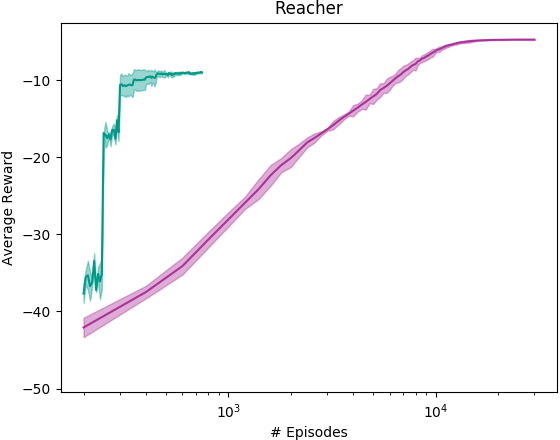}
        \caption{}
    \end{subfigure}
    \caption{(a)~Comparison of our method to PPO on the 2D navigation task presented in the paper. Our method uses roughly three orders of magnitude fewer samples to solve the task compared to PPO. (b)~On the car from images task, our method achieves slightly worse performance than PPO though with about 25 times fewer samples. (c)~Comparison of our method to PPO for the reacher task. Our method achieves worse final performance but uses about 40 times fewer samples than these methods.}
    \label{fig:log-plots}
\end{figure}

In \autoref{fig:log-plots} we include the plots for the simulated tasks comparing \metabbr\ and PPO. Note that the x-axis is on a log scale, i.e., though our method is sometimes worse in final policy performance, we use one to three orders of magnitude fewer samples. This demonstrates our method's sample efficiency compared to PPO, while being able to solve complex image-based domains that are difficult for model-based methods.

PPO is an on-policy model-free RL method, and typically off-policy methods exhibit better sample efficiency \citep{td3,sac}. We use PPO in our comparisons because on-policy methods are typically easier to tune, at the cost of being less efficient, and the complexity of our image-based environments poses a major challenge for all RL methods. Specifically, we also compared to TD3 \citep{td3}, and we were unable to train successful policies despite extensive hyperparameter tuning. We also note that, to our knowledge, TD3 has never been tested on image-based domains.

